\documentclass[10pt,twocolumn,letterpaper]{article}

\usepackage[pagenumbers]{cvpr} 
\usepackage{url}
\usepackage{booktabs}       
\usepackage{amsfonts}       
\usepackage{nicefrac}       
\usepackage{microtype}      
\usepackage{multirow}
\usepackage{pifont}
\usepackage{amsmath}
\usepackage{graphicx}
\usepackage{floatrow}
\usepackage{caption}

\definecolor{cvprblue}{rgb}{0.21,0.49,0.74}
\usepackage[pagebackref,breaklinks,colorlinks,allcolors=cvprblue]{hyperref}

\def\paperID{xxxx} 
\def\confName{CVPR}
\def\confYear{2025}

\title{Relation3D: Enhancing Relation Modeling for Point Cloud Instance Segmentation}

\author{Jiahao Lu$^1$ \quad Jiacheng Deng$^1$\thanks{Corresponding Author} \\
$^1$University of Science and Technology of China
}
\begin{document}
\maketitle

\renewcommand{\thefootnote}{\fnsymbol{footnote}}
\vspace{-2em}
\begin{abstract}
    3D instance segmentation aims to predict a set of object instances in a scene, representing them as binary foreground masks with corresponding semantic labels. Currently, transformer-based methods are gaining increasing attention due to their elegant pipelines and superior predictions. However, these methods primarily focus on modeling the external relationships between scene features and query features through mask attention. They lack effective modeling of the internal relationships among scene features as well as between query features.
In light of these disadvantages, we propose \textbf{Relation3D: Enhancing Relation Modeling for Point Cloud Instance Segmentation}. Specifically, we introduce an adaptive superpoint aggregation module and a contrastive learning-guided superpoint refinement module to better represent superpoint features (scene features) and leverage contrastive learning to guide the updates of these features.
Furthermore, our relation-aware self-attention mechanism enhances the capabilities of modeling relationships between queries by incorporating positional and geometric relationships into the self-attention mechanism.
Extensive experiments on the ScanNetV2, ScanNet++, ScanNet200 and S3DIS datasets demonstrate the superior performance of Relation3D. Code is available at \href{https://github.com/Howard-coder191/Relation3D}{\color{cyan}this website}.
 
\end{abstract}

\section{Introduction}

Point cloud instance segmentation aims to identify and segment multiple instances of specific object categories in 3D space. With the rapid development of fields such as robotic grasping~\cite{zhuang2023instance}, augmented reality~\cite{park2020deep,manni2021snap2cad}, 3D/4D reconstruction~\cite{wu20244d,zhu2024motiongs,lu2024align3r,lu2024dn,luiten2023dynamic}, and autonomous driving~\cite{neven2018towards,yurtsever2020survey}, as well as the widespread application of LiDAR and depth sensor technologies~\cite{lehtola2017comparison,halber2019rescan}, point cloud instance segmentation has become a core technology for achieving efficient and accurate scene understanding. However, the unordered, sparse, and irregular nature of point cloud data, combined with the complex distribution of objects and the numerous categories~\cite{li2025pamba,deng2024diff3detr} in real-world applications, presents unique challenges for effective point cloud analysis. To tackle these challenges, early approaches mainly concentrated on accurately generating 3D bounding boxes (top-down)~\cite{yi2019gspn,hou20193d,yang2019learning} or effectively grouping the points into instances with clustering algorithms (bottom-up)~\cite{engelmann20203d,liang2021instance,vu2022softgroup}. However, these methods have limitations: they either heavily rely on high-quality 3D bounding boxes, or require manual selection of geometric attributes.
\begin{figure}[!t]
    \begin{center}\includegraphics[width=1\textwidth]{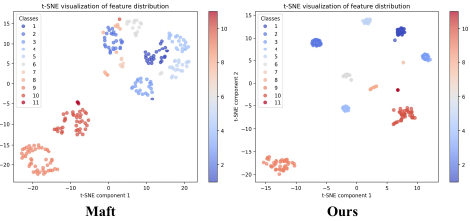}
        \caption{\textbf{T-SNE visualization of the superpoint-level feature distributions on ScanNetV2 validation set.} Different colors represent different instances. Our method highlights better inter-object diversity and intra-object similarity.
        }
        \vspace{-1.em}
        \label{T-SNE}
    \end{center}
\end{figure}
\begin{table}[!tbp]

\caption{\textbf{Excessive feature variation among points within the same supepoint and comparison of $L_{cont}$ in different settings.} The experiment is conducted on ScanNetV2 validation set. $L_{cont}$ measures the consistency of superpoint features within the same instance and the differences between features of different instances. Detailed information about $L_{cont}$ can be found in Equation~\ref{cont}. Stage 1 represents the features output by ASAM, while stages 2 and 3 represent the features after refinement by CLSR. \vspace{-1.8em}}%
  \label{table:intro}
  \centering
  \scalebox{0.7}{\begin{tabular}{c|cccc}
      \toprule
      \toprule
       \multicolumn{3}{c|}{Point feature variation (Maft) }& \multicolumn{2}{c}{1.8603 (Excessive) }\\
      \midrule
      \midrule
      Setting&Maft & Ours (Stage 1) & Ours (Stage 2)   & Ours (Stage 3)  \\
      \midrule
      $L_{cont}$&1.057&0.7255&0.5841&\textbf{0.5739}\\
      \bottomrule
  \end{tabular}}
\end{table}

Recently, researchers start to focus on the design of transformer-based methods~\cite{schult2022mask3d,sun2023superpoint,lu2023query,lai2023mask, lu2025beyond}. These methods adopt an encoder-decoder framework (end-to-end), where each object instance is represented by an instance query. The encoder is responsible for learning the point cloud scene features, while the transformer decoder~\cite{cheng2021per} iteratively attends to the point cloud scene features to learn the instance queries. Ultimately, the instance queries can directly generate the masks for all instances in parallel. %
Current mainstream transformer-based methods commonly use mask-attention~\cite{cheng2022masked} to effectively model the external relationships between scene features and query features. However, they lack effective modeling for the internal relationships among scene features and between query features. As shown in Table~\ref{table:intro} and the left panel of Figure~\ref{T-SNE}, we observe insufficient consistency in superpoint features within the same instances, inadequate differentiation between features of different instances, and excessive feature variation among points within the same superpoint. These erroneous relationships between scene features undoubtedly increase the difficulty of instance segmentation. Besides, the effectiveness of self-attention lies in its establishment of relationships between query features. However, simply computing similarity between query features is too implicit and lacks adequate spatial and geometric relationship modeling, whose importance has been demonstrated in~\cite{lin2021core,hou2024relation,hao2023relation}. Although position embeddings are used to guide self-attention in transformer-based methods, the spatial information position embeddings provide is typically imprecise. For instance, SPFormer's~\cite{sun2023superpoint} position embeddings are learnable and lack concrete spatial meaning, and in methods like Mask3D~\cite{schult2022mask3d}, Maft~\cite{lai2023mask}, and QueryFormer~\cite{lu2023query}, discrepancies exist between the positions indicated by position embeddings and the actual spatial locations of each query's corresponding mask. This limitation in conventional self-attention prevents effective integration of implicit relationship modeling with spatial and geometric relationship modeling.

Based on the above discussion, we summarize two core issues that need to be considered and addressed in point cloud instance segmentation: \textit{1) How to effectively model the relationships between scene features?} Most previous methods use pooling operations to obtain superpoint features~\cite{sun2023superpoint,lai2023mask,li2025sas}, but this pooling operation introduces unsuitable features and blurs distinctive features when there are large feature differences between points within a superpoint. Therefore, we need a new way to model superpoint features to  emphasizing the distinctive point features. Additionally, considering the significant feature differences between superpoints within the same instance, we need to introduce scene feature relation priors to guide the superpoint features and model better superpoint relationships. \textit{2) How to better model the relationships between queries?} Current self-attention designs rely on a simple computation of similarity between queries, but this implicit relationship modeling often requires extensive data and prolonged training to capture meaningful information. Therefore, integrating explicit spatial and geometric relationships is crucial, as it can refine attention focus areas and accelerate convergence. This motivates us to introduce instance-related biases to enhance the modeling of spatial and geometric relationships effectively.

Inspired by the above discussion, we propose \emph{Relation3D: Enhancing Relation Modeling for Point Cloud Instance Segmentation}, which includes an \underline{a}daptive \underline{s}uperpoint \underline{a}ggregation \underline{m}odule (ASAM), a \underline{c}ontrastive \underline{l}earning-guided \underline{s}uperpoint \underline{r}efinement module (CLSR), and \underline{r}elation-aware \underline{s}elf-\underline{a}ttention (RSA).
To address the first issue, we propose an adaptive superpoint aggregation module and a contrastive learning-guided superpoint refinement module. In the adaptive superpoint aggregation module, we adaptively calculate weights for all points within each superpoint, emphasizing distinctive point features while diminishing the influence of unsuitable features.  In the contrastive learning-guided superpoint refinement module, we first adopt a dual-path structure in the decoder, with bidirectional interaction and alternating updates between query features and superpoint features. This design enhances the representation ability of superpoint features. Furthermore, to optimize the update direction of superpoint features, we introduce contrastive learning~\cite{khosla2020supervised,chen2020simple,lu2024bsnet} to provide contrastive supervision for superpoint features, reinforcing the consistency of superpoint features within instances and the differences between features of different instances, as is shown in Figure~\ref{T-SNE}. To address the second issue, in relation-aware self-attention, we first model the explicit relationships between queries. By obtaining the mask and its bounding box corresponding to each query, we can model the positional and geometric relationships between queries. Next, we embed these relationships into self-attention as embeddings. Through this approach, we achieve an effective integration of implicit relationship modeling with spatial and geometric relationship modeling.

The main contributions of this paper are as follows: (i) We propose Relation3D: Enhancing Relation Modeling for Point Cloud Instance Segmentation, which achieves accurate and efficient point cloud instance segmentation predictions. (ii) The adaptive superpoint aggregation module and the contrastive learning-guided superpoint refinement module effectively enhances the consistency of superpoint features within instances and the differences between features of different instances. The relation-aware self-attention improves the relationship modeling capability between queries by incorporating the positional and geometric relationships into self-attention. (iii) Extensive experimental results on four standard benchmarks, ScanNetV2~\cite{dai2017scannet}, ScanNet++~\cite{yeshwanth2023scannet++}, ScanNet200~\cite{rozenberszki2022language}, and S3DIS~\cite{armeni20163d}, show that our proposed model achieves superior performance compared to other transformer-based methods.

\section{Related Work}

{\bf Proposal-based Methods.}
Existing proposal-based methods are heavily influenced by the success of Mask R-CNN~\cite{he2017mask} for 2D instance segmentation. The core idea of these methods is to first extract 3D bounding boxes and then use a mask learning branch to predict the mask of each object within the boxes. GSPN~\cite{yi2019gspn} adopts an analysis-by-synthesis strategy to generate high-quality 3D proposals, refined by a region-based PointNet~\cite{qi2017pointnet}. 3D-BoNet~\cite{yang2019learning} employs PointNet++\cite{qi2017pointnet++} for feature extraction from point clouds and applies Hungarian Matching\cite{kuhn1955hungarian} to generate 3D bounding boxes. These methods set high expectations for proposal quality.

\begin{figure*}[!ht]
    \vspace{-2em}
    \begin{center}
        \includegraphics[width=0.9\textwidth]{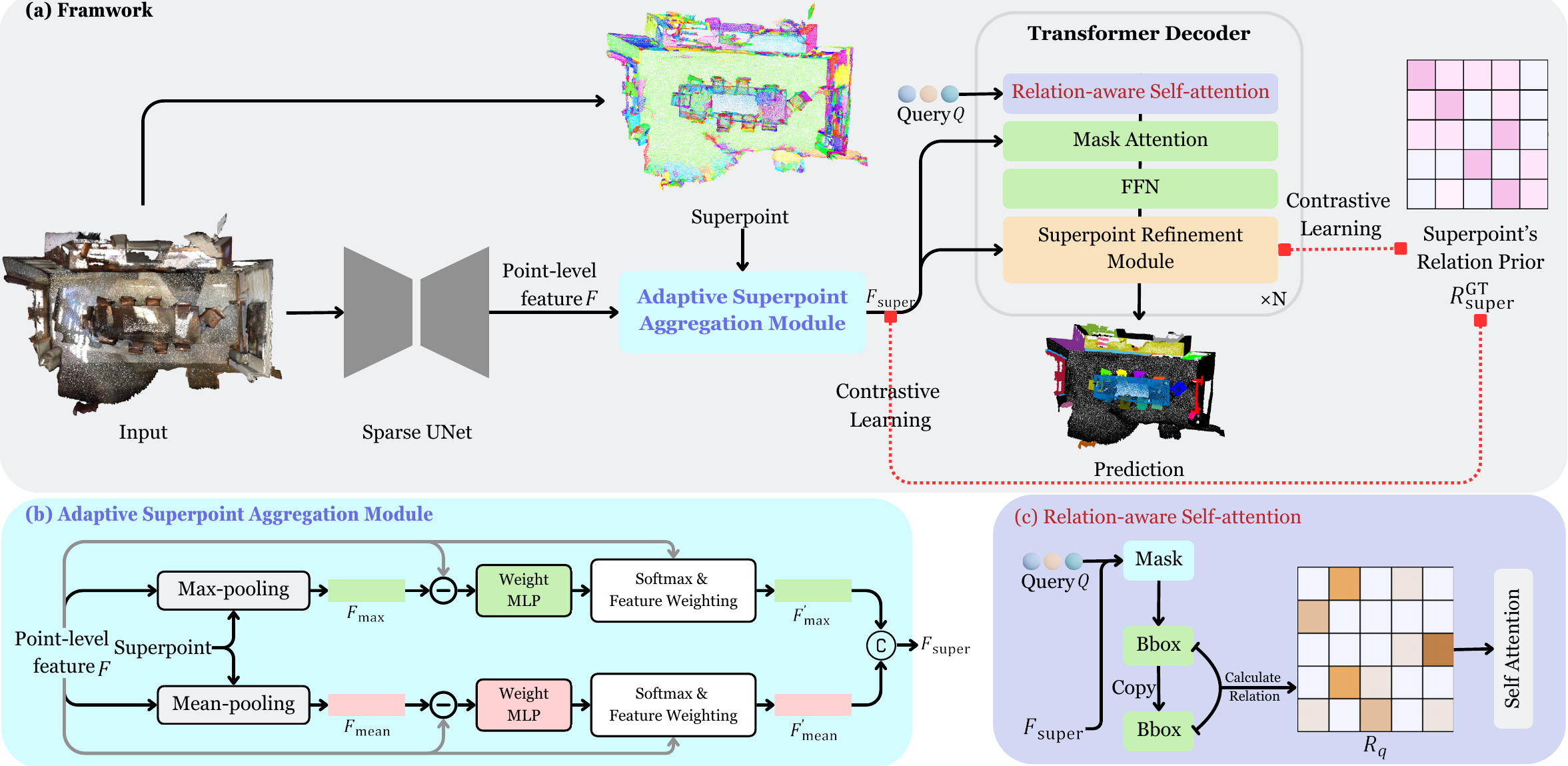}
        \caption{(a) The overall framework of our method Relation3D. (b) The details of our proposed adaptive superpoint aggregation module. (c) The details of our proposed relation-aware self-attention.\vspace{-1.0em}
        }
        \label{framework}
    \end{center}
    \vspace{-1.em}
\end{figure*}
{\bf Grouping-based Methods.}
Grouping-based methods follow a bottom-up processing flow, first generating predictions for each point (such as semantic mapping and geometric displacement), and then grouping the points into instances based on these predicted attributes. 
PointGroup~\cite{jiang2020pointgroup} segments objects on original and offset-shifted point clouds and employs ScoreNet for instance score prediction. %
SoftGroup~\cite{vu2022softgroup} groups on soft semantic scores and uses a top-down refinement stage to refine the positive samples and suppress false positives.
ISBNet~\cite{ngo2023isbnet} introduces a cluster-free approach utilizing instance-wise kernels. Recently, Spherical Mask~\cite{shin2024spherical} has addressed the low-quality outcomes of coarse-to-fine strategies by introducing a new alternative instance representation based on spherical coordinates.

{\bf Transformer-based Methods.}
Following 2D instance segmentation techniques~\cite{zheng2021rethinking, cheng2021per, li2023mask}, in the 3D field, each object instance is represented as an instance query, with query features learned through a vanilla transformer decoder. Transformer-based methods require the encoder to finely encode the point cloud structure in complex scenes and use the attention mechanism in the decoder to continuously update the features of the instance queries, aiming to learn the complete structure of the foreground objects as much as possible. Mask3D~\cite{schult2022mask3d} and SPFormer~\cite{sun2023superpoint} are pioneering works utilizing the transformer framework for 3D instance segmentation, employing FPS and learnable queries, respectively, for query initialization. QueryFormer~\cite{lu2023query} and Maft~\cite{lai2023mask} build on Mask3D~\cite{schult2022mask3d} and SPFormer~\cite{sun2023superpoint} by improving query distribution. However, these works have not thoroughly explored the importance of internal relationships between scene features and between query features. Our method aims to enhance relation modeling for both scene features and query features to achieve better instance segmentation.

{\bf Relation modeling.}
Many 2D methods have demonstrated the importance of relation modeling. CORE~\cite{lin2021core} first leverages a vanilla relation block to model the relations among all text proposals and further enhances relational reasoning through instance-level sub-text discrimination in a contrastive manner. RE-DETR~\cite{hao2023relation} incorporates relation modeling into component detection by introducing a learnable relation matrix to model class correlations. Relation-DETR~\cite{hou2024relation} explores incorporating positional relation priors as attention biases to augment object detection. Our method is the first to explore the significance of relation priors in 3D instance segmentation. Through the adaptive superpoint aggregation module and the contrastive learning-guided superpoint refinement module, we progressively enhance the relationships among scene features. Additionally, relation-aware self-attention improves the relationships among queries. 

\section{Method}
\subsection{Overview}
The goal of 3D instance segmentation is to determine the categories and binary masks of all foreground objects in the scene. 
The architecture of our method is illustrated in Figure~\ref{framework}.
Assuming that the input point cloud
has $N$ points, each point contains position $(x, y, z)$, color
$(r, g, b)$ and normal $(n_x, n_y, n_z)$ information.
Initially, we utilize a Sparse UNet~\cite{spconv2022} to extract point-level feature $F \in \mathbb{R} ^{N\times C}$. Next, we perform adaptive superpoint aggregation module (Section~\ref{asam}) to acquire the superpoint-level features $F_{\rm super}\in \mathbb{R} ^{M\times C}$. Subsequently, we initialize several instance queries $Q \in \mathbb{R} ^{K\times C}$ and input $Q$ and $F_{\rm super}$ into the transformer decoder. To improve the relationship modeling capability between queries, we propose the relation-aware self-attention (Section~\ref{rsa}). 
To update the features of $ F_{\rm super} $, we design a superpoint refinement module (Section~\ref{clsr}) in the decoder, which is also a cross attention operation. However, unlike conventional cross-attention, the scene features $ F_{\rm super} $ act as the $\mathcal{Q}$, while the instance queries $Q$ serve as the $\mathcal{K}$ and $\mathcal{V}$. To guide the update direction of the superpoint features $ F_{\rm super} $, we implement a contrastive learning approach, which enhances the consistency of superpoint features within instances and increases the differences between features of different instances.

\subsection{Backbone}
We employ Sparse UNet~\cite{spconv2022} as the backbone for feature extraction, yielding features $F$, which is consistent with SPFormer~\cite{sun2023superpoint} and Maft~\cite{lai2023mask}. Next, we aggregate the point-level features $F$ into superpoint-level features $F_{\rm super} $ via adaptive superpoint aggregation module, which will be introduced in the subsequent section. 
\subsection{Adaptive Superpoint Aggregation Module}
\label{asam}
The purpose of this module is to aggregate point-level features into superpoint-level features. To emphasize distinctive and meaningful point features while diminishing the influence of unsuitable features, we design the adaptive superpoint aggregation module, as shown in Figure~\ref{framework} (b). Specifically, we first perform max-pooling and mean-pooling on the point-level features $F$ according to the pre-obtained superpoints, resulting in $F_{\rm max}$ and $F_{\rm mean}$ respectively. Next, we calculate the difference between the superpoint-level features and the original point-level features $F$. We then utilize two non-shared weight MLPs to predict the corresponding weights,
\begin{gather}
  \label{weights}
  \mathcal{W}_{\rm max} = {\rm MLP_1}(F_{\rm max}-F),\\
  \mathcal{W}_{\rm mean} = {\rm MLP_2}(F_{\rm mean}-F).
\end{gather}
Getting the corresponding weights $ \mathcal{W}_{\rm max} $ and $ \mathcal{W}_{\rm mean} $, we apply a softmax operation to them in each superpoint. In this way, we can obtain the contribution of each point to its corresponding superpoint. We then use these weights, which sum to 1 in each superpoint, to perform feature weighting on $ F $, resulting in $ F_{\rm max}' $ and $ F_{\rm mean}' $. It's worth noting that the computation for each superpoint can be parallelized with point-wise MLP and torch-scatter extension library~\cite{paszke2019pytorch}, so this superpoint-level aggregation is actually efficient. Finally, we concatenate $ F_{\rm max}' $ and $F_{\rm mean}' $ to $[F'_{\rm max}, F'_{\rm min}]$ and input them into an MLP to reduce the $2C$ channels to $C$, obtaining the final superpoint-level features $F_{\rm super}\in \mathbb{R} ^{M\times C}$. 

\subsection{Contrastive Learning-guided Superpoint Refinement Module}
\label{clsr}
In the previous section, we introduce the adaptive superpoint aggregation module to emphasize distinctive point features within superpoints. Next, to further enhance the expressiveness of superpoints, we will leverage query features to update superpoint features within the transformer decoder. This design, in conjunction with the original mask attention, forms a dual-path architecture, enabling direct communication between query and superpoint features. This approach accelerates the convergence speed of the iterative updates. 

\begin{figure}[!ht]
    \begin{center}\includegraphics[width=0.8\textwidth]{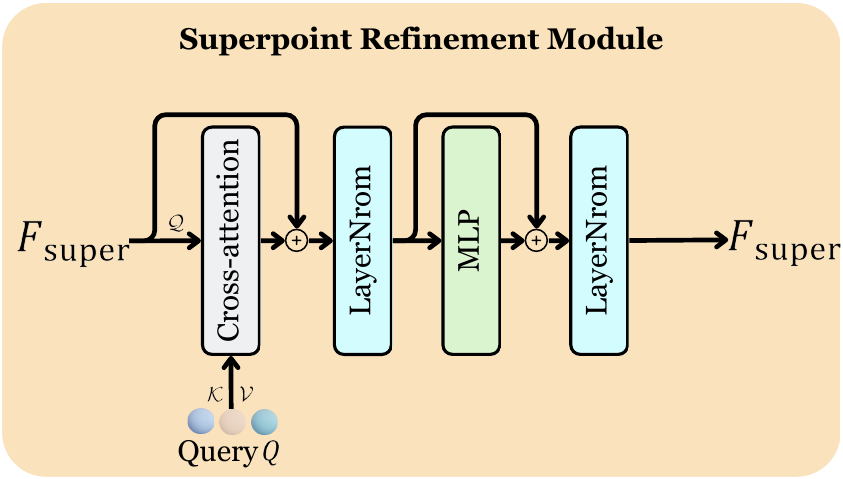}
        \caption{The superpoint refinement module. Superpoint-level features \( F_{\rm super} \) serve as the \( \mathcal{Q} \) in cross-attention, while the instance queries \( Q \) serve as the \( \mathcal{K} \) and \( \mathcal{V} \). \vspace{-1.em}}
        \vspace{-2.em}
        \label{sfm}
    \end{center}
\end{figure}
Specifically, the superpoint refinement module employs a cross-attention mechanism for feature interaction. Here, we use the superpoint-level features $ F_{\rm super} $ as the $\mathcal{Q}$ in the cross attention, while the instance queries $ Q $ serve as the $\mathcal{K}$ and $\mathcal{V}$. The specific structure is illustrated in Figure~\ref{sfm}. To reduce computational and memory costs, we do not perform self-attention for self-updating $ F_{\rm super} $. Furthermore, the superpoint refinement module is not applied at every decoder layer. Instead, we perform the refinement of $ F_{\rm super} $ every $ r $ layers to reduce computational resource consumption .

Furthermore, to guide the update direction, we have designed a contrastive learning mechanism, which constrains the consistency of superpoint features within the same instance and enlarge the differences between features of different instances
First, we can obtain the superpoint's relation prior $ R_{\rm super}^{\rm GT} $ based on instance annotations. If the current scene has $ M $ superpoints, then $ R_{\rm super}^{\rm GT} $ is an $ M \times M $ binary matrix defined as follows,
\begin{equation}
R_{\rm super}^{\rm GT}(i,j) = \begin{cases}
1, & \text{if } i \text{ and } j \text{ are in the same instance;} \\
0, & \text{otherwise,}
\end{cases}
\end{equation}
where $ i $ and $ j $ represent two different superpoints.
Next, we will compute the similarity between  $ F_{\rm super} $ features, defined as follows,
\begin{equation}
\mathcal{S}  = {\mathbf{Norm} }(F_{\rm super})@{\mathbf{Norm} }(F_{\rm super})^{\rm T}.
\end{equation}
Here, $\mathcal{S}$ represents the similarity matrix where each element quantifies the relationship between pairs of superpoint features. 
Finally, we will apply contrastive learning by comparing $\mathcal{S}$ and $R_{\rm super}^{\rm GT}$ as follows,
\begin{equation}
\label{cont}
L_{cont}  = \text{BCE}(\frac{\mathcal{S}+1}{2}, R_{\rm super}^{\rm GT}),
\end{equation}
where $L_{cont}$ is the contrastive loss computed using binary cross-entropy (BCE). This loss will encourage the model to enhance the consistency of superpoint features within the same instance while reinforcing the differences between features of different instances. Notably, we also add this loss function after the adaptive superpoint aggregation module, which can guide ASAM to focus on meaningful features within the superpoint that help enhance the consistency of superpoint features within the same instance.
\subsection{Relation-aware Self-attention}
\label{rsa}
Previous methods use traditional self-attention to model the relationships between queries, where each query contains a content embedding and a position embedding. They first add the position embedding to the content embedding before computing the attention map. However, in most methods~\cite{schult2022mask3d,sun2023superpoint,lu2023query,lai2023mask}, the position embedding does not accurately match the actual position of the mask predicted by the corresponding query, leading to imprecise implicit modeling of positional relationships. More explanation can be found in the supplemental materials. Inspired by Relation-DETR~\cite{hou2024relation}, to enhance the self-attention's ability to model positional relationships and to improve geometric relationship modeling, we propose a relation-aware self-attention (RSA). 

To be specific, we calculate the binary mask $\mathbb{M}$ for each instance query $Q$. 
Next, we calculate the bounding box (bbox) corresponding to each mask, including its center point and scale: $ x, y, z, l, w, h $. With the bbox calculated, we compute the relative relationships between queries as follows,

\textbf{i. Positional Relative Relationship:}
\begin{footnotesize}
   \[
   \left[ \log\left(\frac{|x_i - x_j|}{l_i} + 1\right), \log\left(\frac{|y_i - y_j|}{w_i} + 1\right), \log\left(\frac{|z_i - z_j|}{h_i} + 1\right) \right];
   \]
\end{footnotesize}

\textbf{ii. Geometric Relative Relationship:}
   \[
   \left[ \log\left(\frac{l_i}{l_j}\right), \log\left(\frac{w_i}{w_j}\right), \log\left(\frac{h_i}{h_j}\right) \right],
   \]
where $i, j$ represents two different queries.
Next, we concatenate these two sets of relationships to form an embedding, denoted as $ \mathfrak{T} \in \mathbb{R}^{K \times K \times 6} $. Then, following past methods, we use conventional sine-cosine encoding to increase the dimensionality of $ \mathfrak{T}\in \mathbb{R}^{K \times K \times 6d} $,
\begin{equation}
  \mathfrak{T}' = \sin\cos(\mathfrak{T}).
\end{equation}
Finally, the embedding $ \mathfrak{T}'$ undergoes a linear transformation to obtain $R_q \in \mathbb{R}^{K \times K \times \mathcal{H}}$, where $\mathcal{H}$ denotes the number of attention heads.

After obtaining $R_q$, we incorporate it into the traditional self-attention mechanism. The specific formula is as follows,
\begin{equation}
  {\rm RSA}(Q) = {\rm Softmax}(\frac{\mathcal{Q}\mathcal{K}^T}{\sqrt{\mathcal{C}}} +R_q ) \mathcal{V}.
\end{equation}
In this formulation, we have $ \mathcal{Q} = QW_q $, $ \mathcal{K} = QW_k $, and $ \mathcal{V} = QW_v $, where $ W_q $, $ W_k $, and $ W_v $ denote the linear transformation matrices for query, key, and value respectively. 

\subsection{Model Training and Inference}
\label{training inference}
Apart from Maft's losses~\cite{lai2023mask}, our method includes an additional contrastive loss $L_{cont}$, 
\begin{equation}
\begin{aligned}
  \label{lall}
  L_{all} = \lambda_1L_{ce} + \lambda_2L_{bce} + \lambda_3L_{dice}  \\+\lambda_4L_{center} + \lambda_5L_{score}+\lambda_6L_{cont}, 
\end{aligned}
\end{equation}
where $\lambda_1$, $\lambda_2$, $\lambda_3$, $\lambda_4$, $\lambda_5$, $\lambda_6$ are hyperparameters. 
During the model inference phase, we use the predictions from the final layer as the final output. In addition to the normal forward pass through the network, we also employ NMS~\cite{neubeck2006efficient} on the final output as a post-processing operation.

\section{Experiments}
\subsection{Experimental Setup}
\textbf{Datasets and Metrics.}
We conduct our experiments on ScanNetV2~\cite{dai2017scannet}, ScanNet++~\cite{yeshwanth2023scannet++}, ScanNet200~\cite{rozenberszki2022language}, and S3DIS~\cite{armeni20163d} datasets.
\textbf{ScanNetV2} comprises 1,613 scenes with 18 instance categories, of which 1,201 scenes are used for training, 312 for validation, and 100 for testing. \textbf{ScanNet++} contains 460 high-resolution (sub-millimeter) indoor scenes with dense instance annotations across 84 unique instance categories. \textbf{ScanNet200} uses the same point cloud data, but it enhances annotation diversity, covering 200 classes, 198 of which are instance classes. 
\textbf{S3DIS} is a large-scale indoor dataset collected from six different areas, containing 272 scenes with 13 instance categories. Following previous works~\cite{lai2023mask}, we use the scenes in Area 5 for validation and the remaining areas for training.
AP@25 and AP@50 represent the average precision scores with IoU thresholds of 25\% and 50\%, respectively. mAP is the mean of all AP scores, calculated with IoU thresholds ranging from 50\% to 95\% in 5\% increments.
On ScanNetV2, we report mAP, AP@50, and AP@25. Additionally, we report Box AP@50 and AP@25 results, as done in SoftGroup~\cite{vu2022softgroup} and Maft~\cite{lai2023mask}. For ScanNet200 and ScanNet++, we report mAP, AP@50, and AP@25. On S3DIS, we report AP@50 and AP@25.

\textbf{Implementation Details.}
\label{Implementation}
We build our model on PyTorch framework~\cite{paszke2019pytorch} and train our model on a single RTX4090 with a batch size of 6 for 512 epochs. We employ Maft~\cite{lai2023mask} as the baseline. We employ AdamW~\cite{loshchilov2017decoupled} as the optimizer and PolyLR as the scheduler, with a maximum learning rate of 0.0002.
Point clouds are voxelized with a size of 0.02m.
For hyperparameters, we tune $K, r$ as 400, 3 respectively.
$\lambda_1 ,\lambda_2 ,\lambda_3 ,\lambda_4 ,\lambda_5, \lambda_6$ in Equation~\ref{lall} are set as 0.5, 1, 1, 0.5, 0.5, 1. Since ScanNet++ and ScanNet200 have more categories and instances, we set $K$ as 500. All the other hyperparameters are the same for all datasets. %
\subsection{Comparison with existing methods.}
\begin{figure}[!t]
    \begin{center}
        \includegraphics[width=1\textwidth]{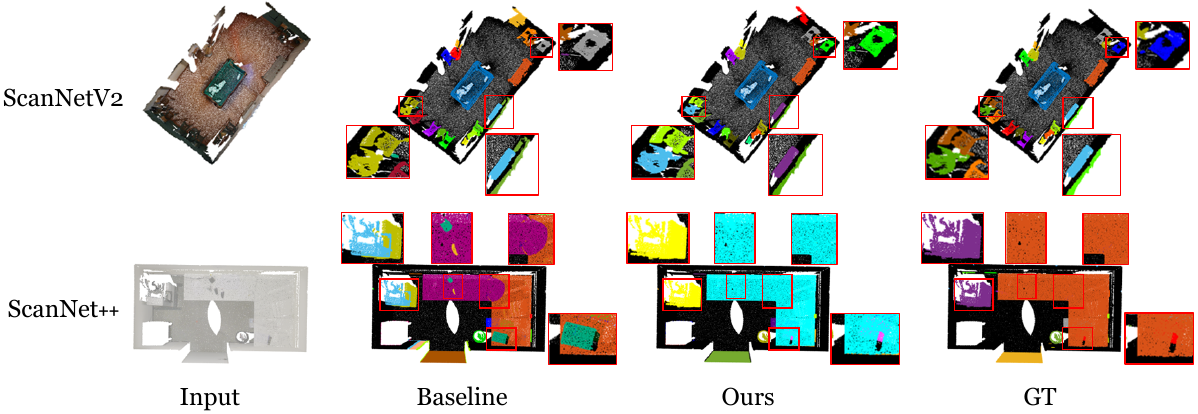}
        \caption{\textbf{Visualization of instance segmentation results on ScanNetV2 and ScanNet++ validation set. }The red boxes highlight the key regions.}
        \label{compare}
    \end{center}
    \vspace{-1.em}
\end{figure}

\textbf{Results on ScanNetV2.}
Table~\ref{table:ScanNetV2} reports the results on ScanNetV2 validation and hidden test set.
Due to our focus on modeling the internal relationships between the scene features and between the queries, our approach outperforms other transformer-based methods, achieving an increase in mAP by 2.6, AP@50 by 3.7, AP@25 by 2.5, Box AP@50 by 2.8 and Box AP@25 by 1.8 in the validation set, and a rise in mAP by 2.6, AP@50 by 3.0 and AP@25 by 4.1 in the hidden test set. To vividly illustrate the differences between our method and others, we visualize the qualitative results in Figure~\ref{compare}. From the regions highlighted in red boxes, it is evident that our method can generate more accurate predictions.

\begin{table}[!t]
  \begin{center}
    \footnotesize
    \setlength\tabcolsep{1.5pt}
    \caption{\textbf{Comparison on ScanNetV2 validation and hidden test set.} The second and third rows are the non-transformer-based and transformer-based methods, respectively. $\ddagger$ denotes using surface normal.}
    \label{table:ScanNetV2}
    \scalebox{0.76}{\begin{tabular}{c|ccccc|ccc}
      \toprule
      \multirow{2}*{Method} &  \multicolumn{5}{c|}{ScanNetV2 validation} &  \multicolumn{3}{c}{ScanNetV2 test}\\
       & mAP & AP@50 & AP@25 & Box AP@50 & Box AP@25 & mAP & AP@50 &AP@25\\
      \midrule
      3D-SIS~\cite{hou20193d}  & / & 18.7 & 35.7 & 22.5 & 40.2 &16.1 &38.2&55.8 \\
      3D-MPA~\cite{engelmann20203d}     & 35.3 & 51.9 & 72.4 & 49.2 & 64.2 &35.5 &61.1&73.7 \\
      DyCo3D~\cite{he2021dyco3d}     & 40.6 & 61.0 & / & 45.3 & 58.9 &39.5 &64.1&76.1 \\
      PointGroup~\cite{jiang2020pointgroup}         & 34.8 & 56.9  & 71.3 & 48.9 & 61.5  &40.7 &63.6&77.8 \\
      MaskGroup~\cite{zhong2022maskgroup}         & 42.0 &  63.3  & 74.0 & / & / &43.4 & 66.4&79.2 \\
      OccuSeg~\cite{han2020occuseg}         &  44.2 & 60.7  & / & / & /  &48.6 &67.2&74.2 \\
      HAIS~\cite{chen2021hierarchical}                     & 43.5 & 64.4  & 75.6 & 53.1 & 64.3 &45.7 &69.9&80.3 \\
      SSTNet~\cite{liang2021instance}                     & 49.4 & 64.3  & 74 & 52.7 & 62.5 &50.6 &69.8&78.9	\\
      SoftGroup~\cite{vu2022softgroup}                     & 45.8 & 67.6  & 78.9 & 59.4 & 71.6 &50.4 &76.1&86.5 \\
      DKNet~\cite{wu20223d}                     & 50.8 & 66.9  & 76.9 & 59.0 & 67.4&53.2&71.8&81.5 \\   
      ISBNet~\cite{ngo2023isbnet} & 54.5 &73.1 &82.5 & 62.0 &78.1&55.9 &75.7&83.5 \\   
      Spherical Mask~\cite{shin2024spherical} &\textbf{62.3} &\textbf{79.9} &\textbf{88.2}&/&/&\textbf{61.6}& \textbf{81.2} &\textbf{87.5} \\
      \midrule
      Mask3D~\cite{schult2022mask3d}            & 55.2 & 73.7  & 82.9 & 56.6 & 71.0 &56.6& 78.0&87.0\\
      QueryFormer~\cite{lu2023query}     & 56.5 & 74.2  & 83.3 &61.7& 73.4& 58.3 &78.7&87.4 \\
      SPFormer~\cite{sun2023superpoint}     & 56.3 & 73.9  & 82.9 &/& / & 54.9 &77.0&85.1\\
      Maft~\cite{lai2023mask}     & 58.4 & 75.9  & 84.5 &63.9& 73.5 &57.8 &77.4 &/\\
      Maft$\ddagger$~\cite{lai2023mask}     & 59.9 &76.5  & / &/& / &59.6 &78.6&86.0\\
      Ours  &\textbf{62.5 } & \textbf{80.2 } &  \textbf{87.0} & \textbf{66.7}&  \textbf{75.3} & \textbf{62.2}&\textbf{81.6} & \textbf{90.1}\\
      \bottomrule
    \end{tabular}}
     \vspace{-1.em}
  \end{center}
\end{table}
\textbf{Results on ScanNet++.}
Table~\ref{table:ScanNetpp} presents the results on ScanNet++ validation and hidden test set. The notable performance enhancement underscores the efficacy of our method in handling denser point cloud scenes.
\begin{table}[!t]
  \begin{center}
    \footnotesize
    \setlength\tabcolsep{3pt}
 
   \caption{\textbf{Comparison on ScanNet++ validation and hidden test set.} ScanNet++ contains denser point cloud scenes and wider instance classes than ScanNetV2, with 84 distinct instance classes.}
    \label{table:ScanNetpp}
    \begin{tabular}{c|ccc|ccc}
    \toprule 
    \multirow{2}*{Method} &  \multicolumn{3}{c|}{ScanNet++ validation} &  \multicolumn{3}{c}{ScanNet++ test}\\
    & mAP & AP@50 & AP@25 & mAP & AP@50 & AP@25\\
    \midrule
    PointGroup~\cite{jiang2020pointgroup}  & /&/&/&8.9&14.6 &21.0 \\
    HAIS~\cite{chen2021hierarchical} & /&/&/ & 12.1&19.9 &29.5 \\
    SoftGroup~\cite{vu2022softgroup}&/ & /&/&16.7&29.7 &38.9 \\
    Maft~\cite{lai2023mask}	&23.1	&32.6	&39.7&20.9	&31.3	&40.4\\
    Ours &\textbf{28.2}	&\textbf{39.3}	&\textbf{46.1}& \textbf{24.2} & \textbf{35.5} &\textbf{44.0} \\
    \bottomrule
  \end{tabular}
     \vspace{-1.em}
  \end{center}
\end{table}

\textbf{Results on ScanNet200.}
Table~\ref{table:ScanNet200} reports the results on ScanNet200 validation set. The significant performance improvement demonstrates the effectiveness of our method in handling complex and challenging scenes with a broader range of categories. 
\begin{table}[!tbp]
\caption{\textbf{Comparison on ScanNet200 validation set.} ScanNet200 employs the same point cloud data as ScanNetV2 but enhances more annotation diversity, with 198 instance classes.}
  \label{table:ScanNet200}
  \centering
  \scalebox{0.8}{    
  \begin{tabular}{c|ccc}
    \toprule 
    \multirow{2}*{Method} &  \multicolumn{3}{c}{ScanNet200 validation} \\
    & mAP & AP@50 & AP@25 \\
    \midrule
    SPFormer~\cite{sun2023superpoint} & 25.2&33.8 &39.6 \\
    Mask3D~\cite{schult2022mask3d} & 27.4&37.0 &42.3\\
    QueryFormer~\cite{lu2023query} & 28.1&37.1 &43.4 \\
    Maft~\cite{lai2023mask} &29.2 & 38.2 &43.3 \\
    Ours & \textbf{31.6 } & \textbf{41.2 } &\textbf{45.6 } \\
    \bottomrule
  \end{tabular}}
\end{table}

\begin{table}[!tbp]
\caption{\textbf{Comparison on S3DIS Area5.} S3DIS contains 13 instance
categories. }%
  \label{table:S3DIS}
  \centering
    \scalebox{0.8}{\begin{tabular}{c|cc}
      \toprule 
      Method & AP@50& AP@25 \\
      \midrule
      PointGroup~\cite{jiang2020pointgroup} & 57.8 &/ \\
      MaskGroup~\cite{zhong2022maskgroup} & 65.0 &/ \\
      SoftGroup~\cite{vu2022softgroup} & 66.1 &/ \\
      SSTNet~\cite{liang2021instance} & 59.3 &/ \\
      SPFormer~\cite{sun2023superpoint} & 66.8 &/ \\
      Mask3D~\cite{schult2022mask3d} & 68.4 &75.2 \\
      QueryFormer~\cite{lu2023query} & 69.9 & /\\
      Maft~\cite{lai2023mask} & 69.1& 75.7 \\
      Spherical Mask~\cite{shin2024spherical} &72.3 &/\\
      Ours & \textbf{72.5}&\textbf{78.5}\\
      \bottomrule
    \end{tabular}}

\end{table}

\textbf{Results on S3DIS.}
We evaluate our method on S3DIS using Area 5 in Table~\ref{table:S3DIS}. Our proposed method achieves better performance compared to previous methods, with gains in both AP@50 and AP@25, demonstrating the effectiveness and generalization of our method.

\subsection{Ablation Studies}

\begin{table}[!tbp]
\caption{\textbf{Evaluation of the model with different designs on ScanNetV2 validation set.} ASAM refers to the adaptive superpoint aggregation module. CLSR refers to the contrastive learning-guided superpoint refinement module. RSA refers to the relation-aware self-attention. \vspace{-1em}}%
  \centering
    \label{table:ablation}
  \scalebox{0.8}{\begin{tabular}{c|ccc|ccc}
      \toprule
      &ASAM & CLSR & RSA  & mAP & AP@50 &AP@25\\
      \midrule
      {[A]} &\ding{55}&\ding{55}&\ding{55}& 59.8&77.4&85.4  \\
     
      {[B]} &\ding{51}&\ding{55}&\ding{55}& 60.1&77.9&85.6	\\

      {[C]} &\ding{55}&\ding{51}&\ding{55}&60.9&78.7&86.2\\
      {[D]} &\ding{51}&\ding{51}&\ding{55}&61.5&78.8&86.3 \\
      {[E]} &\ding{55}&\ding{55}&\ding{51}&61.0 &78.5 &86.0\\
      {[F]} &\ding{51}&\ding{51}&\ding{51}	&\textbf{62.5}&\textbf{80.2} &\textbf{87.0}\\
      \bottomrule
  \end{tabular}}
\end{table}
\textbf{Evaluation of the model with different designs.}
To further study the effectiveness of our designs, we conduct ablation studies on ScanNetV2 validation set. 
As shown in Table~\ref{table:ablation}, {[A]} represents the baseline of our method, which is Maft~\cite{lai2023mask} using surface normals and NMS. [B] demonstrates that with the assistance of ASAM, which aims to better aggregate point-level features into superpoint-level features and emphasize distinctive and meaningful point features while diminishing the influence of unsuitable features, there is a performance improvement: mAP increases by 0.3, AP@50 by 0.5, and AP@25 by 0.2. However, due to the lack of guidance from contrastive loss (introduced in the CLSR), the aggregation direction of superpoints cannot be effectively controlled as expected, so the performance gain is limited. To validate this point, as shown in Table~\ref{table:Effectiveness}, adding contrastive loss to guide ASAM leads to further performance enhancement. 

[C] shows that with the help of CLSR, we can interactively update superpoint features, and the use of contrastive learning guides the update direction by enforcing consistency of superpoint features within the same instance and increasing the difference between features of different instances. Compared to the baseline [A], the performance improves significantly, with mAP increasing by 1.1, AP@50 by 1.3, and AP@25 by 0.8. {[D]} combines ASAM and CLSR. In this design, not only the contrastive learning embedded within CLSR provides ASAM with a clear direction for feature aggregation but also ASAM can offer better-initialized superpoint features to CLSR. This synergistic design cooperates well and results in a 0.6 mAP improvement. {[E]} demonstrates the effectiveness of RSA, which enhances the self-attention mechanism’s ability to model positional relationships and improves geometric relationship modeling. Compared to [A], RSA leads to an improvement of 1.2 mAP and 1.1 AP@50. Finally, in [F], we present the performance of the complete model, underscoring the essential roles played by each module in 3D instance segmentation.

\textbf{Importance of different designs.} \textcolor{blue}{I:} RSA incorporates explicit relationship modeling between queries, which helps the network learn and converge more easily by focusing on more relevant queries, compared to purely implicit modeling. \textcolor{blue}{II:} ASAM and CLSR are essentially a unified entity (designed to solve the same problem: better modeling of scene relationships). We separated them only for clarity in description. Both \textcolor{blue}{I} and \textcolor{blue}{II} are equally important for instance segmentation, and their contributes comparably to the final performance.

\begin{figure}[!ht]
    \begin{center}
        \includegraphics[width=1\textwidth]{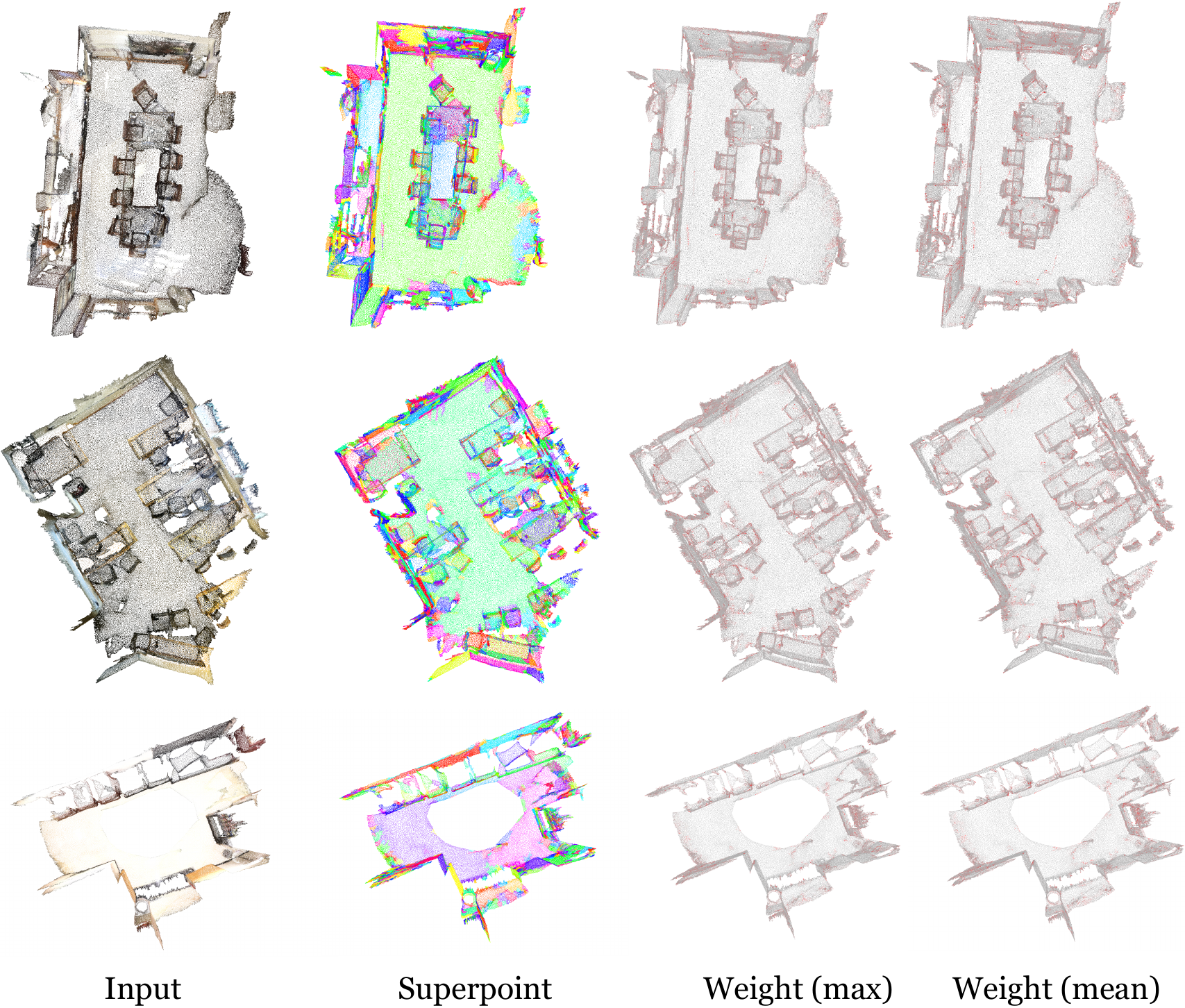}
        \caption{\textbf{Visualization of weights in the adaptive superpoint aggregation module.} A deeper red color indicates a higher weight assigned to point feature during the ``Softmax \& Feature Weighting'' stage (Figure~\ref{framework} (b)).
        }
        \label{weight1}
    \end{center}
    \vspace{-1.em}
\end{figure}

\begin{table}[!tbp]
\caption{\textbf{Effectiveness of contrastive loss to ASAM.} }%
  \label{table:Effectiveness}
  \scalebox{0.8}{\begin{tabular}{c|ccc}
      \toprule
      Setting &  mAP & AP@50 &AP@25\\
      \midrule
        W/o contrastive loss &60.1&77.9&85.6\\
      W contrastive loss&\textbf{60.5}&\textbf{78.4}&\textbf{85.9}\\
      \bottomrule
  \end{tabular}}
\end{table}
\textbf{Ablation study on the adaptive superpoint aggregation module.}
In this section, we conduct experiments on the adaptive superpoint aggregation module (ASAM). First, we perform an ablation study on max-pooling and mean-pooling, as shown in Table~\ref{table:asam}, where ``W max-pooling'' indicates that ASAM includes only the max-pooling branch. The results show that both mean-pooling and max-pooling contribute to performance gains. Furthermore, to illustrate the characteristics of the learned weight distribution in ASAM, we present corresponding visualization in Figure~\ref{weight1}. From the figure, it is evident that ASAM places greater emphasis on the edges and corner regions of objects—areas that are typically distinctive for each instance. Therefore, with the assistance of ASAM and contrastive learning, the model is able to aggregate more discriminative superpoint features.

\begin{table}[!tbp]
\caption{\textbf{Ablation study on ASAM.}  \vspace{-1em}}%
  \label{table:asam}
  \scalebox{0.8}{\begin{tabular}{c|ccc}
      \toprule
      Setting &  mAP & AP@50 &AP@25\\
      \midrule
      W max-pooling &62.2&79.9&86.7\\
      W mean-pooling&62.3&79.7&86.7\\
      W max-pooling \& mean-pooling&\textbf{62.5}&\textbf{80.2}&\textbf{87.0}\\
      \bottomrule
  \end{tabular}}
\end{table}

\begin{figure*}[!ht]
    \vspace{-2em}
    \begin{center}
        \includegraphics[width=1\textwidth]{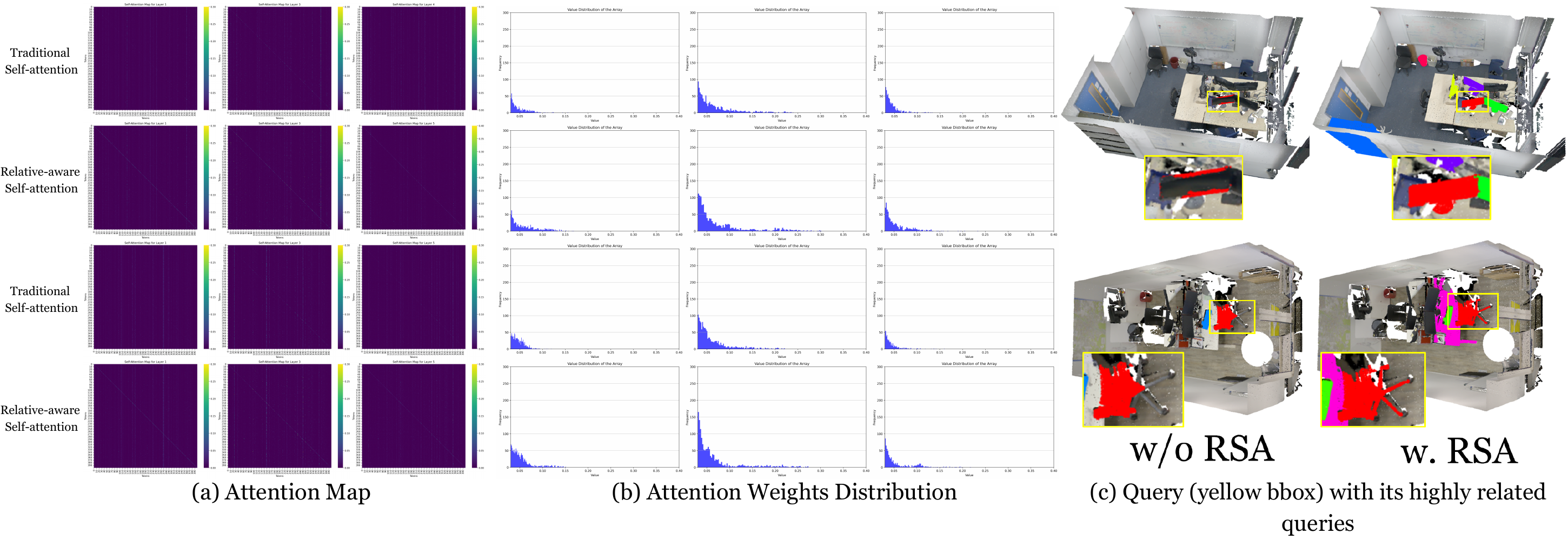}
        \caption{\textbf{(a) Comparison of attention maps for traditional self-attention vs. relation-aware self-attention.} We display the progression of attention maps from layer 1, 3, 5. \textbf{(b) Comparison of attention weight distributions for traditional self-attention vs. relation-aware self-attention.} The attention weight distributions are also shown from layer 1, 3, 5. \textbf{(c) Query (yellow bbox) with its highly related queries.} \vspace{-1em}}
        \label{attentionmap1}
    \end{center}
    \vspace{-1.em}
\end{figure*}

\textbf{Effectiveness of the relation-aware self-attention.}
As shown in Figure~\ref{attentionmap1}, we compare the attention maps and attention weight distributions between traditional self-attention and relation-aware self-attention. From Figure~\ref{attentionmap1}(a), it can be observed that our proposed relation-aware self-attention has more high-weight focal points in its attention map, which is further supported by the data in Figure~\ref{attentionmap1}(b). Notably, we have excluded points with attention values ranging from 0 to 0.03 from our statistical analysis, as these account for the vast majority (approximately 99\%) of the attention map and would otherwise obscure the meaningful patterns in our study. Furthermore, we substantiate this from a visualization perspective. As demonstrated in Figure~\ref{attentionmap1}(c), with the aid of RSA, the representative query can forge connections with a broader set of relevant queries, in contrast to traditional attention mechanisms that concentrate on a limited number of specific queries. This enhancement facilitates the generation of superior instance masks. These observations indicate that relation-aware self-attention achieves a more focused attention when modeling position and geometric relationships. Unlike traditional self-attention, which has relatively dispersed attention without any specific focal query, relation-aware self-attention selectively emphasizes relevant queries, resulting in a more precise and meaningful representation.

\begin{figure}[!ht]
    \begin{center}
        \includegraphics[width=1\textwidth]{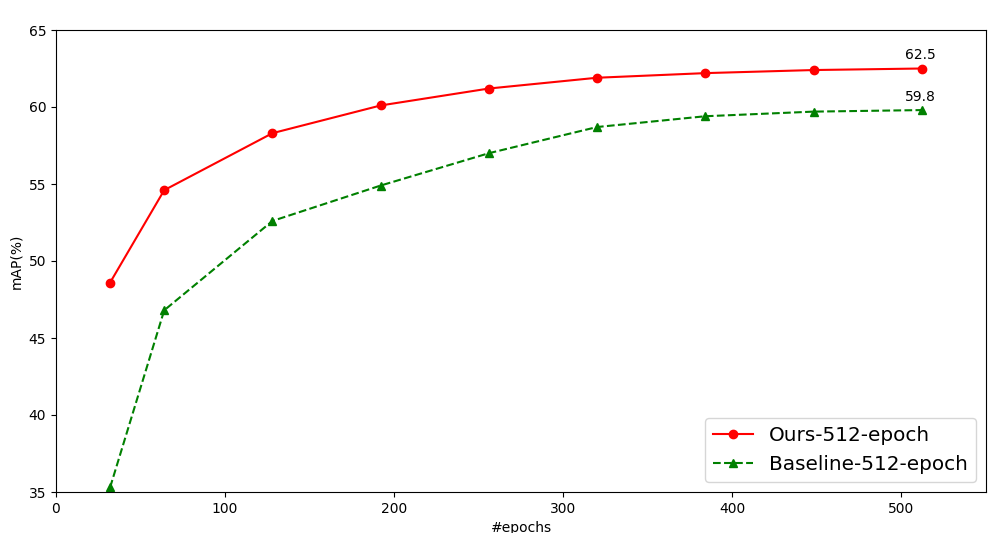}
        \caption{\textbf{The convergence curve on ScanNet-v2 validation set.}}
        \label{curve}
    \end{center}
    \vspace{-2.5em}
\end{figure}

\textbf{Contribution to the convergence speed.}
As shown in Figure~\ref{curve}, our method demonstrates a faster convergence speed compared to the baseline. This improvement can be attributed to the relation priors introduced by CLSR and RSA: contrastive learning provides relation priors for superpoints to guide feature aggregation, while RSA introduces position and geometric relation priors for query features, enhancing self-attention. Additionally, the superpoint refinement module in CLSR forms a dual-path architecture, enabling direct communication between query features and superpoint features, speeding up the convergence.

\section{Conclusion}
In this paper, we propose a novel 3D instance segmentation method called Relation3D. We focus on modeling the internal relationships among scene features as well as between query features, an aspect that past methods have not explored sufficiently. Specifically, we introduce an adaptive superpoint aggregation module to better represent superpoint features and a contrastive learning-guided superpoint refinement module that updates superpoint features in dual directions while guiding the direction of these updates with the help of contrastive learning. Additionally, our proposed relation-aware self-attention mechanism enhances the modeling of relationships between queries by improving the representation of positional and geometric relationships. Extensive experiments conducted on the several datasets demonstrate the effectiveness of Relation3D.

\section{Acknowledgements}
This work was partially supported by Wenfei Yang and Tianzhu Zhang.


{
    \small
    \bibliographystyle{unsrt}
    \bibliography{egbib}
}
\clearpage
\setcounter{page}{1}
\maketitlesupplementary

\section{Overview} %
In this supplementary material, we begin by presenting a more detailed comparison of quantitative metrics on ScanNetV2~\cite{dai2017scannet} validation set and test set (Section~\ref{Detailed_Results}).
We then provide additional discussion about position embedding (Section~\ref{Discussion}).
To further validate the effectiveness of the proposed method, we provide more visualizations (Section~\ref{Visualization}). 

\section{Detailed results on ScanNetV2 validation and hidden test set}
\label{Detailed_Results}
The detailed results for each category on ScanNetV2 validation set are reported in Table~\ref{table:ScanNetV2val}. As the table illustrates, 
our method achieves the best performance in 14 out of 18 categories.
The two of them work together to achieve 14 out of 18 categories. The superior performance demonstrates the effectiveness of our method. 
The detailed results for each category on ScanNetV2 hidden test set are reported in Table~\ref{table:ScanNetV2map}, ~\ref{table:ScanNetV250} and ~\ref{table:ScanNetV225}. As the tables illustrate, 
our method achieves the best performance in 13 out of 18 categories in Table~\ref{table:ScanNetV2map}, 10 out of 18 categories in Table~\ref{table:ScanNetV250} and 12 out of 18 categories in Table~\ref{table:ScanNetV225}.
The superior performance demonstrates the effectiveness of our method. 
\begin{table*}[!h]
    \begin{center}
      \small
      \setlength\tabcolsep{3.0pt}
      \caption{\textbf{Full quantitative results of mAP on ScanNetV2 validation set.} For reference purposes, we show the results of fully supervised methods in gray. Best performance of box supervised methods is in boldface.}
      \label{table:ScanNetV2val}
      \vspace{0.5em}
      \begin{tabular}{c|c|cccccccccccccccccccc}
        \toprule
        Method &  mAP & \rotatebox{90}{bathtub} &\rotatebox{90}{bed} &\rotatebox{90}{bookshe.}	 &\rotatebox{90}{cabinet}&\rotatebox{90}{chair}	&\rotatebox{90}{counter}	&\rotatebox{90}{curtain}&\rotatebox{90}{desk}	&\rotatebox{90}{door}&\rotatebox{90}{other}&\rotatebox{90}{picture}&\rotatebox{90}{frige}&\rotatebox{90}{s. curtain}&\rotatebox{90}{sink}&\rotatebox{90}{sofa}&\rotatebox{90}{table}&\rotatebox{90}{toilet}&\rotatebox{90}{window}\\
        \midrule
        \rowcolor{gray!40} PointGroup~\cite{jiang2020pointgroup} &34.8& 59.7& 37.6 &26.7& 25.3 &71.2 &6.9 &26.6 &14.0 &22.9 &33.9 &20.8 &24.6 &41.6 &29.8 &43.4 &38.5 &75.8 &27.5 \\
        \rowcolor{gray!40} SSTNet~\cite{liang2021instance}&49.4&77.7&56.6&25.8&40.6&81.8&22.5&38.4&28.1&42.9&52.0&40.3&43.8&48.9&54.9&52.6&55.7&92.9&34.3 \\
        \rowcolor{gray!40} SoftGroup~\cite{vu2022softgroup} & 45.8 &  66.6  & 48.4 & 32.4 & 37.7 &72.3&14.3&37.6&27.6&35.2& 42.0&34.2&56.2&56.9&39.6& 47.6& 54.1 & 88.5&33.0  \\
        \rowcolor{gray!40} DKNet~\cite{wu20223d}& 50.8 & 73.7  & 53.7 & 36.2 &42.6&80.7&22.7&35.7&35.1&42.7&46.7&51.9&39.9&57.2&52.7&52.4&54.2&91.3&37.2\\   	
        \rowcolor{gray!40} Mask3D~\cite{schult2022mask3d}  & 55.2 & 78.3  & 54.3 & 43.5 & 47.1&82.9& 35.9&48.7 &37.0&54.3&59.7 &53.3& 47.7&47.4  &55.6&48.7&63.8&94.6&39.9\\
        \rowcolor{gray!40} ISBNet~\cite{ngo2023isbnet}&54.5&76.3&58.0&39.3&47.7&83.1&28.8&41.8&35.9&49.9&53.7&48.6&51.6&66.2&56.8&50.7&60.3&90.7&41.1 \\
        \rowcolor{gray!40} SPFormer~\cite{sun2023superpoint}&56.3&83.7&53.6&31.9&45.0&80.7&38.4&49.7&41.8&52.7&55.6&55.0&57.5&56.4&\textbf{59.7}&51.1&62.8&\textbf{95.5}&41.1 \\
        \rowcolor{gray!40} 
        QueryFormer~\cite{lu2023query} & 56.5& 81.3  & 57.7 & \textbf{45.0} & 47.2& 82.0 &37.2&43.2&43.3&54.5&60.5&52.6&54.1&62.7& 52.4 &49.9&60.5 &94.7 &37.4\\
        \rowcolor{gray!40} 
      Maft~\cite{lai2023mask}&58.4&80.1& 58.1&41.8 &48.3 &82.2&34.4& \textbf{55.1} &44.3& 55.0&57.9&61.6&56.4& 63.7&54.4&53.0 &66.3& 95.3&42.9\\
        \midrule
        Ours&\textbf{62.5}&\textbf{84.8}&\textbf{63.7}&42.2&\textbf{54.0}&\textbf{83.9}&\textbf{49.1}&53.8&\textbf{46.7}&\textbf{60.4}   &\textbf{65.0}&\textbf{62.8}&\textbf{61.3}&\textbf{69.2}&57.6&\textbf{61.2}&\textbf{68.9}&94.3&\textbf{46.7}\\
        \bottomrule
      \end{tabular}
    \end{center}
  \end{table*}

\begin{table*}[!h]
  \begin{center}
    \small
    \setlength\tabcolsep{2.5pt}
    \caption{\textbf{Full quantitative results of mAP on the ScanNetV2 test set. Best performance is in boldface.}}
    \label{table:ScanNetV2map}
    \vspace{0.5em}
    \begin{tabular}{c|c|cccccccccccccccccccc}
      \toprule
      Method &  mAP & \rotatebox{90}{bathtub} &\rotatebox{90}{bed} &\rotatebox{90}{bookshe.}	 &\rotatebox{90}{cabinet}&\rotatebox{90}{chair}	&\rotatebox{90}{counter}	&\rotatebox{90}{curtain}&\rotatebox{90}{desk}	&\rotatebox{90}{door}&\rotatebox{90}{other}&\rotatebox{90}{picture}&\rotatebox{90}{frige}&\rotatebox{90}{s. curtain}&\rotatebox{90}{sink}&\rotatebox{90}{sofa}&\rotatebox{90}{table}&\rotatebox{90}{toilet}&\rotatebox{90}{window}\\
      \midrule
    \rowcolor{gray!40} 
      3D-BoNet~\cite{qi2018frustum}   &25.3& 51.9& 32.4 &25.1& 13.7& 34.5& 3.1& 41.9& 6.9& 16.2& 13.1& 5.2& 20.2& 33.8& 14.7& 30.1& 30.3& 65.1& 17.8 \\
     \rowcolor{gray!40} 
      MTML~\cite{lahoud20193d}        &28.2& 57.7& 38.0& 18.2& 10.7& 43.0& 0.1& 42.2& 5.7 &17.9& 16.2& 7.0& 22.9& 51.1& 16.1 &49.1& 31.3& 65.0& 16.2 \\\rowcolor{gray!40} 
      3D-MPA~\cite{engelmann20203d}   &35.5& 45.7& 48.4& 29.9& 27.7& 59.1& 4.7& 33.2 &21.2& 21.7& 27.8& 19.3& 41.3& 41.0& 19.5& 57.4& 35.2& 84.9& 21.3\\\rowcolor{gray!40} 
      DyCo3D~\cite{he2021dyco3d}      &39.5& 64.2& 51.8& 44.7& 25.9& 66.6& 5.0& 25.1& 16.6& 23.1& 36.2& 23.2& 33.1& 53.5& 22.9& 58.7& 43.8& 85.0 &31.7 \\\rowcolor{gray!40} 
      PE~\cite{zhang2021point}        &39.6 &66.7& 46.7& 44.6& 24.3& 62.4& 2.2& 57.7 &10.6& 21.9 &34.0& 23.9& 48.7& 47.5& 22.5& 54.1& 35.0 &81.8& 27.3\\\rowcolor{gray!40} 
      PointGroup~\cite{jiang2020pointgroup} &40.7& 63.9 &49.6& 41.5& 24.3& 64.5& 2.1 &57.0& 11.4& 21.1& 35.9& 21.7& 42.8& 66.6 &25.6 &56.2& 34.1& 86.0& 29.1 \\\rowcolor{gray!40} 
      MaskGroup~\cite{zhong2022maskgroup}   &43.4& 77.8& 51.6 &47.1& 33.0 &65.8 &2.9& 52.6 &24.9 &25.6 &40.0& 30.9 &38.4&29.6& 36.8 &57.5 &42.5& 87.7& 36.2 \\\rowcolor{gray!40} 
      OccuSeg~\cite{han2020occuseg}         & 48.6 &80.2& 53.6 &42.8 &36.9 &70.2 &20.5 &33.1 &30.1 &37.9 &47.4& 32.7& 43.7 &\textbf{86.2} &48.5 &60.1 &39.4 &84.6 &27.3  \\\rowcolor{gray!40} 
      HAIS~\cite{chen2021hierarchical}      &45.7 &70.4& 56.1 &45.7 &36.4 &67.3 &4.6& 54.7& 19.4& 30.8& 42.6& 28.8& 45.4& 71.1& 26.2& 56.3& 43.4& 88.9& 34.4 \\\rowcolor{gray!40} 
      SSTNet~\cite{liang2021instance}       & 50.6 &73.8& 54.9 &49.7& 31.6& 69.3& 17.8& 37.7& 19.8& 33.0& 46.3& 57.6& 51.5& 85.7& \textbf{49.4}& 63.7& 45.7& 94.3& 29.0 \\\rowcolor{gray!40} 
      SoftGroup~\cite{vu2022softgroup}      &50.4& 66.7 &57.9& 37.2& 38.1& 69.4& 7.2& 67.7& 30.3& 38.7& 53.1& 31.9& 58.2& 75.4& 31.8& 64.3& 49.2& 90.7& 38.8 \\\rowcolor{gray!40} 
      DKNet~\cite{wu20223d}                 &53.2 &81.5& 62.4 &51.7& 37.7& 74.9& 10.7& 50.9& 30.4 &43.7& 47.5& 58.1& 53.9& 77.5& 33.9& 64.0& 50.6& 90.1& 38.5 \\  \rowcolor{gray!40}  	
      Mask3D~\cite{schult2022mask3d}& 56.6& 92.6 &	59.7 &	40.8 &42.0&	73.7&	23.9 &59.8 &	38.6 &	45.8 &	54.9&	56.8&	71.6 &	60.1 &	48.0 &	64.6 &	57.5 &92.2 &	36.4\\\rowcolor{gray!40} 
      QueryFormer~\cite{lu2023query} &58.6 &	92.6 &	70.2 &	39.3 &	\textbf{50.5} &	73.7 &	27.7 &	58.3 &	37.5 &	47.9 &	53.5 &	56.8 &	61.5 &	72.0 &	48.1 &	74.5 &59.2 & 95.8 &	36.1 \\\rowcolor{gray!40} 
      Maft~\cite{lai2023mask}&59.6 &	88.9 &	\textbf{72.1} &	44.8 &	46.0 &	76.8 &	25.1 &	55.8 &	40.8 &	50.4 &	53.9 &	61.6 &	61.8 &	85.8 &	48.2 &	68.4 &	55.1 &	93.1 &	\textbf{45.0} \\
      \midrule
      Ours& \textbf{62.2} &	\textbf{92.6} &71.0 &	\textbf{54.1}&	50.2 &	\textbf{77.2} &	\textbf{31.4} &	\textbf{59.8} &	\textbf{42.5} &	\textbf{50.4} &	\textbf{56.5} &	\textbf{65.0} &	\textbf{71.6} &	80.9 &47.6 &	\textbf{74.7} &	\textbf{61.8} &	\textbf{96.3} &36.4 \\
      \bottomrule
    \end{tabular}
  \end{center}
\end{table*}

\begin{table*}[!h]
  \begin{center}
    \small
    \setlength\tabcolsep{2.5pt}
    \caption{\textbf{Full quantitative results of AP@50 on the ScanNetV2 test set. Best performance is in boldface.}}
    \label{table:ScanNetV250}
    \vspace{0.5em}
    \begin{tabular}{c|c|cccccccccccccccccccc}
      \toprule
      Method &  AP@50 & \rotatebox{90}{bathtub} &\rotatebox{90}{bed} &\rotatebox{90}{bookshe.}	 &\rotatebox{90}{cabinet}&\rotatebox{90}{chair}	&\rotatebox{90}{counter}	&\rotatebox{90}{curtain}&\rotatebox{90}{desk}	&\rotatebox{90}{door}&\rotatebox{90}{other}&\rotatebox{90}{picture}&\rotatebox{90}{frige}&\rotatebox{90}{s. curtain}&\rotatebox{90}{sink}&\rotatebox{90}{sofa}&\rotatebox{90}{table}&\rotatebox{90}{toilet}&\rotatebox{90}{window}\\
      \midrule
        \rowcolor{gray!40} 
      3D-BoNet~\cite{qi2018frustum}   &48.8 &	100.0  & 67.2  &	59.0  &	30.1  &	48.4  &	9.8  &	62.0  &	30.6  &	34.1  &	25.9  &	12.5  &	43.4 &	79.6 &	40.2 &	49.9 &	51.3  &	90.9 &	43.9 \\\rowcolor{gray!40} 
      MTML~\cite{lahoud20193d}        &54.9 &	100.0 &	80.7 &	58.8 &	32.7 &	64.7 &	4 &	81.5 &	18.0 &	41.8 &	36.4 &	18.2 &	44.5 &	100.0 &	44.2 & 68.8 & 57.1 &	100.0 & 39.6 \\\rowcolor{gray!40} 
      3D-MPA~\cite{engelmann20203d}   &61.1 &	100.0 & 83.3 &	76.5 &	52.6 &	75.6 &	13.6 &	58.8 &	47.0 &43.8 &	43.2 &	35.8 &	65.0 &	85.7 &	42.9 &	76.5 &	55.7 &	100.0 &	43.0\\\rowcolor{gray!40} 
      DyCo3D~\cite{he2021dyco3d}      &64.1 &	100.0 & 84.1 &	89.3 &	53.1 &	80.2 &	11.5 &	58.8 &	44.8 &	43.8 &	53.7 &	43.0 &	55.0 &	85.7 &	53.4 &	76.4 &	65.7 &	98.7 &	56.8 \\\rowcolor{gray!40} 
      PE~\cite{zhang2021point}        &64.5  &	100.0  &	77.3  &	79.8  &	53.8  &	78.6  &	8.8  &	79.9  &	35.0  &	43.5  &	54.7  &	54.5  &	64.6  &	93.3  &	56.2  &	76.1  &	55.6  &	99.7  &	50.1\\\rowcolor{gray!40} 
      PointGroup~\cite{jiang2020pointgroup} &63.6 & 100.0 &	76.5 &	62.4 &	50.5 &	79.7 &	11.6 &	69.6 &	38.4 &	44.1 &	55.9 &	47.6 &	59.6 &	100.0 &	66.6 &	75.6 &	55.6 &	99.7 &	51.3 \\\rowcolor{gray!40} 
      MaskGroup~\cite{zhong2022maskgroup}   &66.4  &	100.0 &	82.2 &	76.4 &	61.6 &	81.5 &	13.9 &	69.4 &	59.7 &	45.9 &	56.6 &	59.9 &	60.0 &	51.6 &	71.5 &	81.9 &	63.5 &	100.0 &	60.3 \\\rowcolor{gray!40} 
      OccuSeg~\cite{han2020occuseg}         & 67.2 &	100.0&	75.8 &	68.2 &	57.6 &	84.2 &	47.7 & 50.4 &	52.4 &	56.7 &	58.5 &	45.1 &	55.7 &	100.0 &	75.1 &	79.7 &	56.3 &	100.0 &	46.7\\\rowcolor{gray!40} 
      HAIS~\cite{chen2021hierarchical}      &69.9 &	100.0 &	84.9 &	82.0 &	67.5 &	80.8 &	27.9 &	75.7 &	46.5 &	51.7 &	59.6 &	55.9 &	60.0 &	100.0 &	65.4 &	76.7&	67.6 &	99.4 &	56.0\\\rowcolor{gray!40} 
      SSTNet~\cite{liang2021instance}       & 69.8 &	100.0 &	69.7 &	88.8 &	55.6 &	80.3 &	38.7 &	62.6 &	41.7 &	55.6 &	58.5 &	70.2 &	60.0 &	100.0 &	82.4 & 72.0 &	69.2 &	100.0 &	50.9\\\rowcolor{gray!40} 
      SoftGroup~\cite{vu2022softgroup}      &76.1 &	100.0 &	80.8 &	84.5 & 71.6 &	86.2 &	24.3 &	\textbf{82.4} &	65.5 &	62.0 & 73.4 &	69.9 &	79.1 &	98.1 &	71.6 &	84.4 & 76.9 &	100.0 &	59.4\\\rowcolor{gray!40} 
      DKNet~\cite{wu20223d} &71.8 &	100.0 &	81.4 &	78.2 &	61.9 &	87.2 &	22.4 &	75.1 &	56.9 &	67.7 &	58.5 &	72.4 &	63.3 &	98.1 &	51.5 &	81.9 &	73.6 &	100.0 &	61.7\\   	\rowcolor{gray!40} 
      Mask3D~\cite{schult2022mask3d}& 78.0 &	100.0 &	78.6 &	71.6 &	69.6 &	88.5 &	50.0 &	71.4 &	\textbf{81.0} &	67.2 &	71.5 &	67.9 &	80.9 &	100.0 &	\textbf{83.1} &	83.3 &	78.7 &	100.0 &	60.2\\\rowcolor{gray!40} 
      QueryFormer~\cite{lu2023query} & 78.4	&100.0&	93.3&	60.1&	\textbf{75.4}&	88.5	&56.4&	67.7&	66.6&	66.4	&71.6&	67.9	&\textbf{82.0}	&100.0	&83.0&	89.7&	80.4 &	100.0	&62.2\\\rowcolor{gray!40} 
      Maft~\cite{lai2023mask}&78.6 &	100.0 &89.4 &	80.7 &	69.4 &	89.3 &	48.6 &	67.4 &	74.0 &	\textbf{78.6} &	70.4 &	\textbf{72.7} &	73.9 &	100.0 &	70.7 &	84.9 &	75.6 &	100.0 &	\textbf{68.5} \\ 
      \midrule
      Ours&\textbf{81.6} &	\textbf{100.0} &	\textbf{97.1} &\textbf{90.8} &	74.3 &	\textbf{92.3} &	\textbf{57.3} &	71.4 &	69.5 &	73.4 &	\textbf{74.7}&	72.5 &	80.9&	\textbf{100.0} &	81.4 &	\textbf{89.9}&	\textbf{82.0} &	\textbf{100.0} &	61.0 \\
      \bottomrule
    \end{tabular}
  \end{center}
\end{table*}

\begin{table*}[!h]
  \begin{center}
    \small
    \setlength\tabcolsep{2.5pt}
    \caption{\textbf{Full quantitative results of AP@25 on the ScanNetV2 test set. Best performance is in boldface.}}
    \label{table:ScanNetV225}
    \vspace{0.5em}
    \begin{tabular}{c|c|cccccccccccccccccccc}
      \toprule
      Method &  AP@25 & \rotatebox{90}{bathtub} &\rotatebox{90}{bed} &\rotatebox{90}{bookshe.}	 &\rotatebox{90}{cabinet}&\rotatebox{90}{chair}	&\rotatebox{90}{counter}	&\rotatebox{90}{curtain}&\rotatebox{90}{desk}	&\rotatebox{90}{door}&\rotatebox{90}{other}&\rotatebox{90}{picture}&\rotatebox{90}{frige}&\rotatebox{90}{s. curtain}&\rotatebox{90}{sink}&\rotatebox{90}{sofa}&\rotatebox{90}{table}&\rotatebox{90}{toilet}&\rotatebox{90}{window}\\
      \midrule
\rowcolor{gray!40} 
      3D-BoNet~\cite{qi2018frustum}   &68.7 &	100.0 &	88.7 &	83.6 &	58.7 &	64.3 &	55.0 &	62.0 &	72.4 &	52.2&	50.1 &	24.3 &	51.2 &	100.0 &	75.1 &	80.7 &	66.1 &	90.9 &	61.2 \\\rowcolor{gray!40} 
      MTML~\cite{lahoud20193d}        &73.1  &	100.0 &99.2  &	77.9  &	60.9  &	74.6  &	30.8  &	86.7 &	60.1  &	60.7  &	53.9  &	51.9  &	55.0  &	100.0  &	82.4  &	86.9 &	72.9 &	100.0 &	61.6 \\\rowcolor{gray!40} 
      3D-MPA~\cite{engelmann20203d}   &73.7 &100.0 &	93.3 &	78.5&	79.4 &	83.1 &	27.9 &58.8 &	69.5 &	61.6 &	55.9 &55.6 &	65.0 &	100.0 &	80.9 &	87.5 &	69.6 &	100.0 &	60.8\\\rowcolor{gray!40} 
      DyCo3D~\cite{he2021dyco3d}      &76.1 &	100.0 &	93.5 &	89.3 &	75.2 &	86.3 &	60.0 &	58.8 &	74.2&	64.1 &	63.3 &	54.6 &	55.0 &	85.7 &	78.9 &	85.3 &	76.2 &	98.7 &	69.9 \\\rowcolor{gray!40} 
      PE~\cite{zhang2021point}        &77.6 &100.0 &90.0&	86.0 &	72.8 &	86.9&	40.0 &	85.7 &	77.4 &	56.8 &	70.1 &60.2 &	64.6 &	93.3 &	84.3 &	89.0&	69.1 &	99.7 &	70.9\\\rowcolor{gray!40} 
      PointGroup~\cite{jiang2020pointgroup} &77.8 &	100.0 &	90.0 &	79.8 &	71.5 &	86.3 &	49.3 &	70.6 &	89.5 &	56.9 &	70.1 &	57.6 &	63.9 &	100.0 &88.0 &85.1 &	71.9 &	99.7 &	70.9 \\\rowcolor{gray!40} 
      MaskGroup~\cite{zhong2022maskgroup}   &79.2&	100.0 &	96.8 &	81.2 &	76.6 &	86.4 &	46.0 &	81.5&	88.8 &	59.8 &	65.1 &	63.9 &	60.0 &	91.8 &	94.1 &89.6 &	72.1 &	100.0 &	72.3 \\\rowcolor{gray!40} 
      OccuSeg~\cite{han2020occuseg}         & 74.2  &100.0 &	92.3  &	78.5  &	74.5  &	86.7  &	55.7  &	57.8  &	72.9  &	67.0  &	64.4  &	48.8  &	57.7 &	100.0  &	79.4  &	83.0  &62.0 &	100.0  &55.0\\\rowcolor{gray!40} 
      HAIS~\cite{chen2021hierarchical}      &80.3 &	100.0 &	\textbf{99.4} &	82.0 &	75.9 &	85.5 &	55.4 &	88.2&	82.7 &	61.5 &67.6 &	63.8 &64.6 &	100.0 &	91.2&	79.7 &	76.7 &	99.4 &	72.6 \\\rowcolor{gray!40} 
      SSTNet~\cite{liang2021instance}       & 78.9&	100.0 &84.0 &	88.8 &	71.7 &	83.5 &	71.7 &68.4&	62.7 &	72.4 &	65.2 &	72.7 &	60.0 &	100.0 &	91.2 &	82.2 &	75.7 &	100.0 &	69.1\\\rowcolor{gray!40} 
      SoftGroup~\cite{vu2022softgroup}      &86.5&	100.0 &	96.9 &	86.0 &	86.0 &	91.3 &	55.8 &	\textbf{89.9} &	91.1 &76.0 &	\textbf{82.8} &	73.6 &	80.2 &98.0 &	91.9 &	87.5 &	87.7 &	100.0&	82.0  \\\rowcolor{gray!40} 
      DKNet~\cite{wu20223d} &81.5 &	100.0 &	93.0 &	84.4 &	76.5 &	91.5 &	53.4 &	80.5 &	80.5 &	80.7 &	65.4 &	76.3 &65.0 &100.0 &79.4 &	88.1 &	76.6 &	100.0 &	75.8 \\   	\rowcolor{gray!40} 
      Mask3D~\cite{schult2022mask3d}& 87.0 &	100.0 &	98.5 &	78.2 &	81.8 &	93.8 &	76.0 &	74.9 &	92.3&	87.7 &	76.0&	78.5 &	82.0 &	100.0 &	91.2 &	86.4 &	87.8 &	98.3 &	82.5\\\rowcolor{gray!40} 
      QueryFormer~\cite{lu2023query} &87.3 &	100.0 &	97.8 &	80.9 &	87.6&	93.7 &	70.2 &	74.9 &	88.4 &	87.5 &	75.5 &\textbf{78.5} &	\textbf{83.5} &	100.0 &	91.2 &	91.6&	86.9 &	100.0 &	82.5\\\rowcolor{gray!40} 
        Maft~\cite{lai2023mask}&86.0 &	100.0 &	99.0 &	81.0 &	82.9 &	94.9 &	80.9 &	68.8 &	83.6 &	90.4 &	75.1 &	79.6 &	74.1 &	100.0 &	86.4 &	84.8&	83.7 &	100.0 &	\textbf{82.8} \\
        \midrule
        Ours&\textbf{90.1} &	\textbf{100.0} &	97.8 &	\textbf{92.8} &\textbf{87.9} &	\textbf{96.2} &	\textbf{88.2} &	74.9 &	\textbf{94.7} &	\textbf{91.2} &	80.2&	75.3 &	82.0 &	\textbf{100.0} &	\textbf{98.4}&	\textbf{91.9}&	\textbf{89.4} &\textbf{100.0} &	81.5 \\
      \bottomrule
    \end{tabular}
  \end{center}
\end{table*}

\section{Discussion about position embedding}
\label{Discussion}
In DETR-based methods, query typically consist of two embeddings: a content embedding and a position embedding. In the transformer decoder, the position embedding is added to the  content embedding and then input into the self-attention/cross-attention mechanisms for interaction, ultimately generating a new content embedding. It is important to note that in these methods, the position represented by the position embedding does not accurately correspond to the actual location of the mask predicted by the query. In the following sections, we will analyze several representative methods in detail.

\textbf{SPFormer:} In SPFormer, both the content embedding and position embedding are learnable. Consequently, the position embedding does not carry any explicit spatial meaning.

\textbf{Mask3D:} In Mask3D, the position embedding is derived using FPS (Furthest Point Sampling). First, $N$ points are sampled using FPS, and each sampled point is encoded using Fourier or sin-cos encoding to generate the corresponding position embedding. The content embedding, however, is initialized to all zeros. During the subsequent decoder process, Mask3D employs self-attention and cross-attention mechanisms to update the features of the content embedding. However, through experiments, we observe that the positions of the sampling points in Mask3D do not align with the actual positions of the corresponding predicted instances (there are large differences between them). This indicates that the positional relationships introduced by the position embedding are inaccurate.

\textbf{Maft:} Maft also employs a learnable position embedding, but unlike SPFormer, the position embedding \( P \in [0,1]^{N \times 3} \). The position embedding is resized based on the scale of the input scene as follows:  
\[
\hat{P} = P \cdot (p_{\text{max}} - p_{\text{min}}) + p_{\text{min}},
\]  
where \( p_{\text{max}}, p_{\text{min}} \in \mathbb{R}^3 \) denote the maximum and minimum coordinates of the input scene, respectively. As a result, the position embedding in Maft acquires actual coordinate meanings.  
To correlate the position embedding \(\hat{P}\) more closely with the positions of the corresponding predicted instances, Maft introduces a \( C_{\text{center}} \) term in the Hungarian matching cost matrix, representing the distance between \(\hat{P}\) and the ground truth instance center. Furthermore, Maft updates \(\hat{P}\) layer by layer in the decoder, allowing the matched \(\hat{P}\) to progressively approach the ground truth instance center.  
Despite these design enhancements, where \(\hat{P}\) becomes closer to the actual positions of the corresponding predicted instances, some discrepancies remain. These inaccuracies impact the precision of the positional relationships.

\begin{table}[!tbp]
\caption{\textbf{Parameter and runtime analysis of different methods on ScanNetV2 validation set.}  The runtime is measured on the
    same device.}%
  \label{table:RuntimeAnalysis}
  \centering
    \scalebox{0.75}{
    \begin{tabular}{c|c|cc}
      \toprule
      Method &  Parameter(M) & Runtime(ms) \\
      \midrule
      HAIS~\cite{chen2021hierarchical} & 30.9 & 525 \\
     
      SSTNet~\cite{liang2021instance} & /& 663 \\
      SPFormer~\cite{sun2023superpoint} & 17.6 & 390 \\
      Mask3D~\cite{schult2022mask3d} & 39.6 & 525 \\
      SoftGroup~\cite{vu2022softgroup} & 30.9 &535\\
      Maft~\cite{lai2023mask} & 20.1 & 375 \\
      QueryFormer~\cite{lu2023query} & 42.3 & 443 \\
      Spherical Mask~\cite{shin2024spherical} &30.8& 432\\
      Ours & 22.1 & 394 \\
      \bottomrule
    \end{tabular}}
\end{table}

\section{Parameter and Runtime Analysis.}

Table~\ref{table:RuntimeAnalysis} presents model parameters and runtime per scan for various methods evaluated on ScanNetV2 validation set. For a fair comparison, all runtimes are measured on the same RTX 4090 GPU. Compared to Maft, our method achieves better performance with an additional 2.0M parameters. Although our method is slightly slower due to additional modules, the small-scale design allows our approach to outperform most methods in both speed and parameter efficiency.

\section{More Visualization}
\label{Visualization}
\begin{figure*}[!h]
    \begin{center}
        \includegraphics[width=0.9\textwidth]{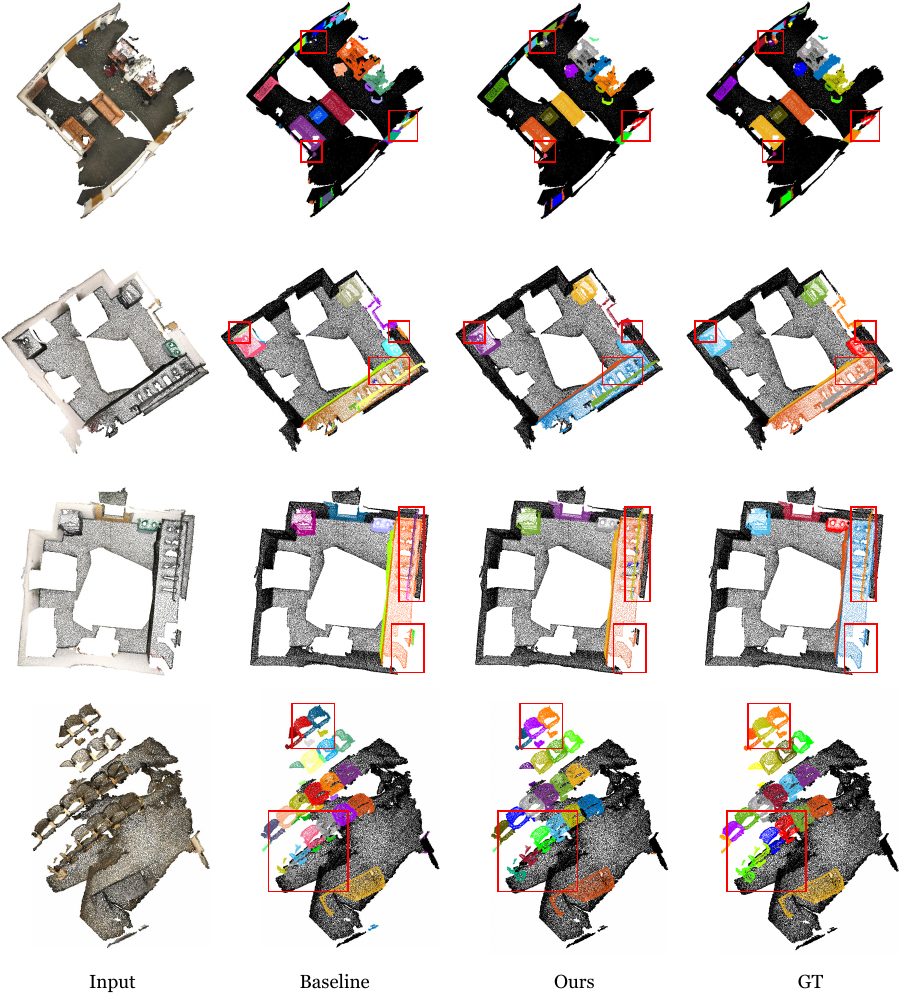}
        \caption{\textbf{More visualization of instance segmentation results on ScanNetV2 validation set.} The red boxes highlight the key regions}
        \label{fig:sup_compare}
    \end{center}
  \end{figure*}

\textbf{Qualitative comparison (Figure~\ref{fig:sup_compare}):}  
To vividly illustrate the differences between our method and baseline, we visualize qualitative results in Figure~\ref{fig:sup_compare}. From the regions highlighted in the last row, we observe that the baseline method tends to confuse chairs with surrounding objects and exhibits incomplete segmentation of the chair. In contrast, our method, by focusing on scene feature modeling, enhances the consistency of superpoint features within instances and increases the differences between features of different instances. This leads to more accurate and coherent segmentation results.

\begin{figure*}[!h]
    \begin{center}
        \includegraphics[width=0.9\textwidth]{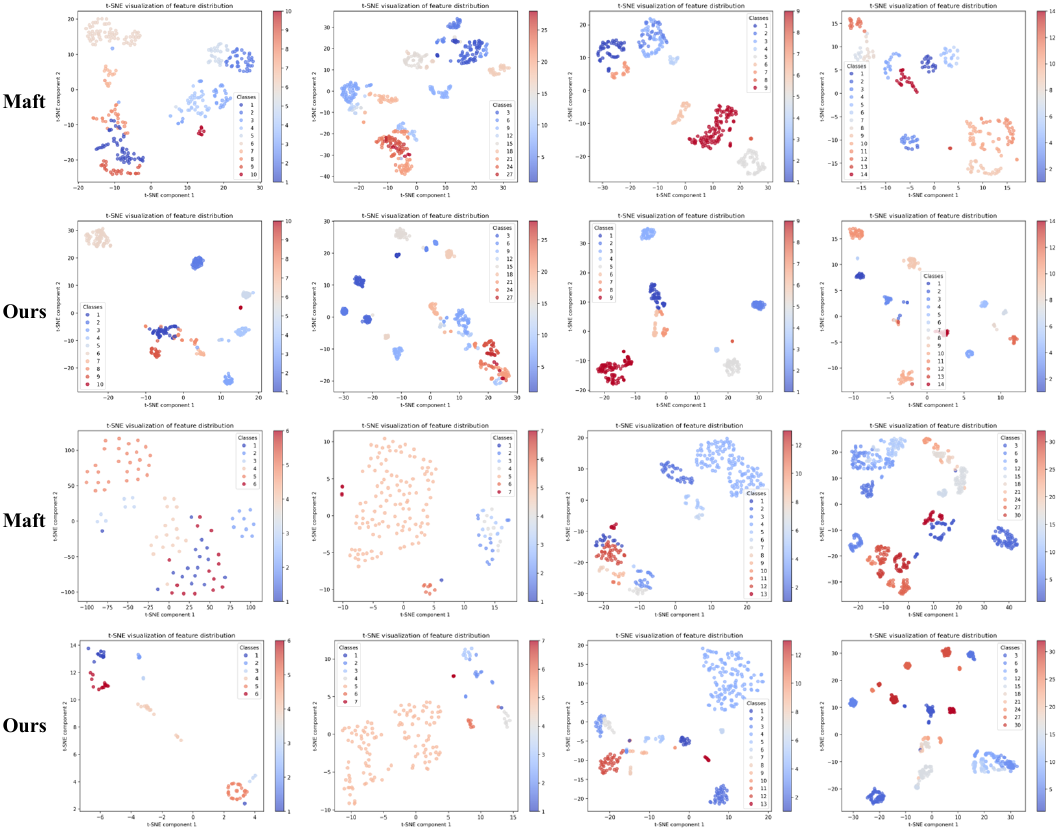}
        \caption{\textbf{More T-SNE visualization of the superpoint-level feature
distributions on ScanNetV2 validation set.}}
        \label{fig:sup_tnse}
    \end{center}
  \end{figure*}

\textbf{T-SNE Visualization (Figure~\ref{fig:sup_tnse}):}  
We present the T-SNE visualization of superpoint-level feature distributions on the ScanNetV2 validation set. From this visualization, it is evident that our method achieves better inter-object differentiation while maintaining intra-object feature consistency. 

\begin{figure*}[!h]
    \vspace{-2em}
    \begin{center}
        \includegraphics[width=0.9\textwidth]{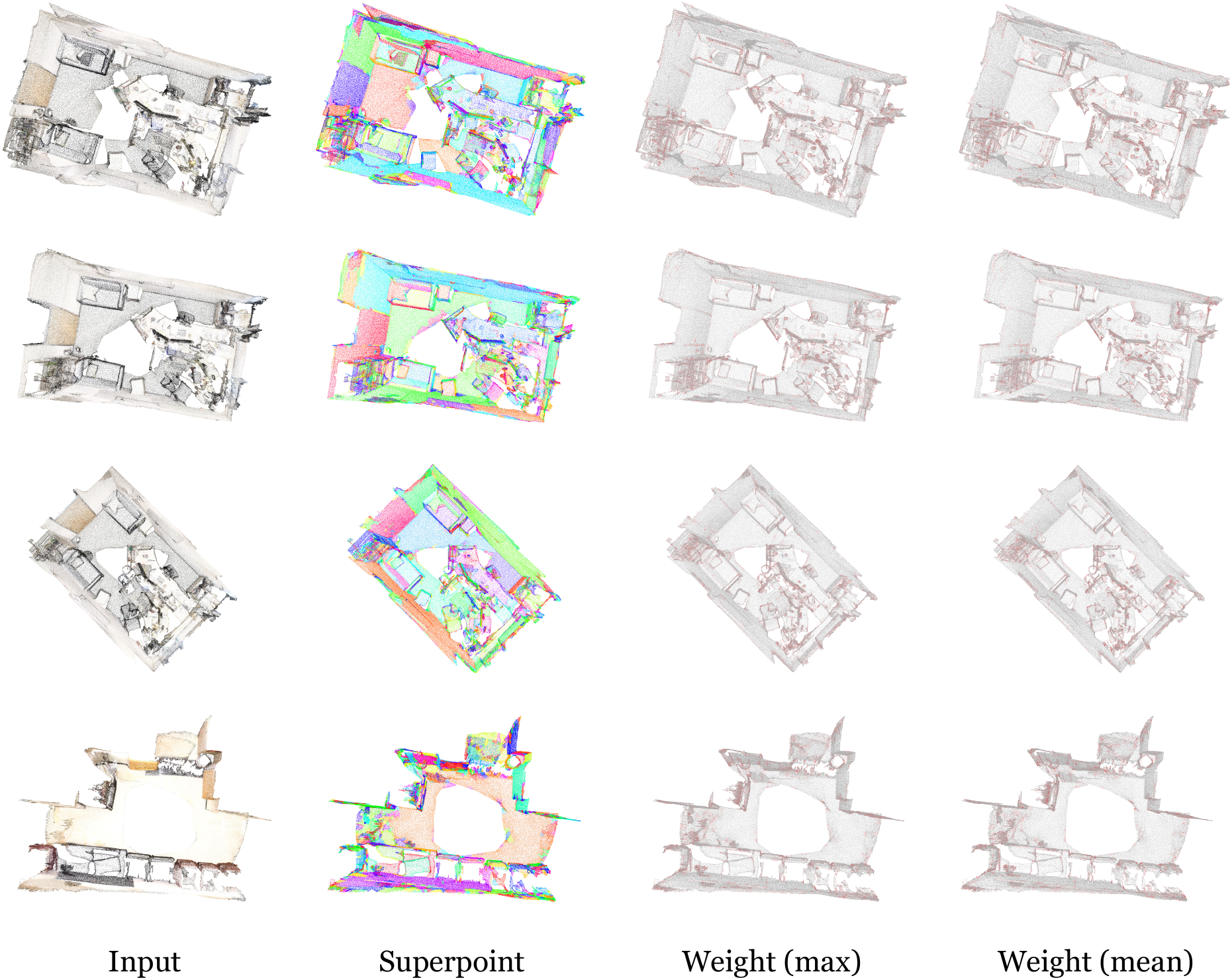}
        \caption{\textbf{More visualization of weights in the adaptive superpoint aggregation module.}
        }
        \label{weight}
    \end{center}
    \vspace{-1em}
  \end{figure*}

\textbf{Adaptive Superpoint Aggregation Module (Figure~\ref{weight}):}  
Figure~\ref{weight} visualizes the weights from the Adaptive Superpoint Aggregation Module (ASAM). The visualization highlights that ASAM assigns higher weights to object edges and corners—regions typically distinctive for individual instances. This targeted emphasis enhances the model's ability to accurately capture key structural features, contributing to better performance in segmentation task.
\section{More ablution study}

We conduct an ablation study to analyze the effect of different values of \( r \) on model performance. By varying \( r \), we evaluate the balance between computational efficiency and the quality of feature refinement. The results of this study are detailed in Table~\ref{table:r}. Through this experiment, we observe that when \( r = 3 \), the performance is comparable to that of \( r = 1 \). However, considering that \( r = 3 \) (the interval of layers after which the refinement of \( F_{\rm super} \) is performed is 3) reduces the frequency of feature refinement, the computational cost is consistently reduced. Thus, we set \( r = 3 \). Moreover, we find that the hyperparameter \( r \) demonstrates strong robustness, consistently achieving a good balance between accuracy and efficiency across different datasets.
\begin{table}[!tbp]
\caption{\textbf{Ablation Study on \( r \).}  
In our method, \( r \) represents the interval of layers after which the refinement of \( F_{\rm super} \) is performed. }%
  \label{table:r}
  \scalebox{0.8}{\begin{tabular}{c|ccc|ccc}
      \toprule
      \multirow{2}*{$r$} &  \multicolumn{3}{c|}{ScanNetV2 validation} &  \multicolumn{3}{c}{ScanNet200 validation}\\
    & mAP & AP@50 & AP@25 & mAP & AP@50 & AP@25\\
      \midrule
      1&62.4&79.9&\textbf{87.1}&31.4&\textbf{41.3}&45.5\\
      3 &\textbf{62.5}&\textbf{80.2} &87.0&\textbf{31.6}& 41.2 &\textbf{45.6}\\
      6 &62.3&79.5 &86.4&30.6& 40.6 &44.9\\
      \bottomrule
  \end{tabular}}
\end{table}
\section{Limitation and future work}
\label{Limitations}
Existing indoor 3D instance segmentation methods primarily focus on static objects and are typically performed offline (i.e., by first reconstructing point clouds from multiple RGB-D frames and then performing segmentation), which is not suitable for embodied environments. Therefore, future work needs to focus more on online instance segmentation in dynamic environments, where reconstruction and segmentation are not decoupled but rather performed simultaneously within a unified framework. A recent work, EmbodiedSAM~\cite{xu2024embodiedsam}, provides valuable insights that are worth exploring.



\end{document}